\documentclass[11pt]{article}
\usepackage[T1]{fontenc}
\usepackage[utf8]{inputenc}
\usepackage{geometry}
\usepackage{color}
\usepackage{mathrsfs}
\usepackage{enumitem}
\usepackage{amsmath}
\usepackage{amsthm}
\usepackage{amssymb}
\usepackage{stmaryrd}
\usepackage[unicode=true,
 bookmarks=false,
 breaklinks=false,pdfborder={0 0 0},pdfborderstyle={},backref=section,colorlinks=true]
 {hyperref}
\hypersetup{
 pdfborderstyle={},linkcolor=blue, citecolor=blue}

\makeatletter
\theoremstyle{plain}
\newtheorem{assumption}{\protect\assumptionname}
\theoremstyle{plain}
\newtheorem{thm}{\protect\theoremname}
\theoremstyle{plain}
\newtheorem{cor}[thm]{\protect\corollaryname}

\allowdisplaybreaks[3]

\makeatother

\providecommand{\assumptionname}{Assumption}
\providecommand{\corollaryname}{Corollary}
\providecommand{\theoremname}{Theorem}

\begin{document}
\title{A Note on the Global Convergence of Multilayer Neural Networks in
the Mean Field Regime}
\author{Huy Tuan Pham\thanks{Department of Mathematics, Stanford University.}$\quad$and
Phan-Minh Nguyen\thanks{Department of Electrical Engineering, Stanford University.}
\thanks{The author ordering is randomized.}}
\maketitle
\begin{abstract}
In a recent work, we introduced a rigorous framework to describe the
mean field limit of the gradient-based learning dynamics of multilayer
neural networks, based on the idea of a neuronal embedding. There
we also proved a global convergence guarantee for three-layer (as
well as two-layer) networks using this framework.

In this companion note, we point out that the insights in our previous
work can be readily extended to prove a global convergence guarantee
for multilayer networks of any depths. Unlike our previous three-layer
global convergence guarantee that assumes i.i.d. initializations,
our present result applies to a type of correlated initialization.
This initialization allows to, at any finite training time, propagate
a certain universal approximation property through the depth of the
neural network. To achieve this effect, we introduce a bidirectional
diversity condition.
\end{abstract}

\section{Introduction}

The mean field (MF) regime refers to a newly discovered scaling regime,
in which as the width tends to infinity, the behavior of an appropriately
scaled neural network under training converges to a well-defined and
nonlinear dynamical limit. The MF limit has been investigated for
two-layer networks \cite{mei2018mean,chizat2018,rotskoff2018neural,sirignano2018mean}
as well as multilayer setups \cite{nguyen2019mean,araujo2019mean,sirignano2019mean,nguyen2020rigorous}.

In a recent work \cite{nguyen2020rigorous}, we introduced a framework
to describe the MF limit of multilayer neural networks under training
and proved a connection between a large-width network and its MF limit.
Underlying this framework is the idea of a neuronal embedding that
encapsulates neural networks of arbitrary sizes. Using this framework,
we showed in \cite{nguyen2020rigorous} global convergence guarantees
for two-layer and three-layer neural networks. It is worth noting
that although these global convergence results were proven in the
context of independent and identically distributed (i.i.d.) initializations,
the framework is not restricted to initializations of this type. In
\cite{nguyen2020rigorous}, it was also proven that when there are
more than three layers, i.i.d. initializations (with zero initial
biases) can cause a certain strong simplifying effect, which we believe
to be undesirable in general. This clarifies a phenomenon that was
first discovered in \cite{araujo2019mean}.

The present note complements our previous work \cite{nguyen2020rigorous}.
Our main task here is to show that the approach in \cite{nguyen2020rigorous}
can be readily extended to prove a similar global convergence guarantee
for neural networks of any number of layers. We however do not assume
i.i.d. initializations. Our result applies to a type of correlated
initialization and the analysis crucially relies on the `neuronal
embedding' framework. As such, our result realizes the vision in \cite{nguyen2019mean}
of a MF limit that does not exhibit the aforementioned simplifying
effect. Furthermore our result cannot be established by the formulations
in \cite{araujo2019mean,sirignano2019mean} which are specific to
i.i.d. initializations.

Similar to the global convergence guarantees in \cite{nguyen2020rigorous}
and unlike other works, our result does not rely critically on convexity
and instead emphasizes on certain universal approximation properties
of neural networks. To be precise, the key is a diversity condition,
which is shown to hold at any finite training time. The insight on
diversity first appeared in the work \cite{chizat2018}: in the context
of two-layer networks, it refers to the full support condition of
the first layer's weight in the Euclidean space. Our previous work
\cite{nguyen2020rigorous} partially hinged on the same insight to
analyze three-layer networks. Here our present result defines a new
notion of diversity in the context of general multilayer networks.
Firstly, it is realized in function spaces that are naturally described
by the `neuronal embedding' framework. Secondly, it is bidirectional:
roughly speaking, for intermediate layers, diversity holds in both
the forward and backward passes. The effect of bidirectional diversity
is that a certain universal approximation property, at any finite
training time, is propagated from the first layer to the second last
one.

\paragraph*{Organization.}

We first describe the multilayer setup and the MF limit in Section
\ref{sec:setup} to make the note self-contained. Our main result
of global convergence (Theorem \ref{thm:global-optimum}) is presented
and proven in Section \ref{sec:global_conv}. This result is proven
for the MF limit. Lastly Section \ref{sec:connection} connects the
result to large-width multilayer networks.

Since the emphasis here is on the global convergence result, to keep
the note concise, other results are stated with proofs omitted, since
they can be found or established in a similar manner to \cite{nguyen2020rigorous}.

\paragraph*{Notations.}

We use $K$ to denote a generic constant that may change from line
to line. We use $\left|\cdot\right|$ to denote the absolute value
for a scalar, the Euclidean norm for a vector, and the respective
norm for an element of a Banach space. For an integer $n$, we let
$\left[n\right]=\left\{ 1,...,n\right\} $. We write ${\rm cl}\left(S\right)$
to denote the closure of a set $S$ in a topological space.

\section{Multilayer neural networks and the mean field limit\label{sec:setup}}

\subsection{Multilayer neural network\label{subsec:Three-layer-neural-network}}

We consider the following $L$-layer network:
\begin{align}
\hat{{\bf y}}\left(x;\mathbf{W}\left(k\right)\right) & =\varphi_{L}\left(\mathbf{H}_{L}\left(x,1;\mathbf{W}\left(k\right)\right)\right),\label{eq:three-layer-nn}\\
\mathbf{H}_{i}\left(x,j_{i};\mathbf{W}\left(k\right)\right) & =\frac{1}{n_{i-1}}\sum_{j_{i-1}=1}^{n_{i-1}}{\bf w}_{i}\left(k,j_{i-1},j_{i}\right)\varphi_{i-1}\left({\bf H}_{i-1}\left(x,j_{i-1};\mathbf{W}\left(k\right)\right)\right),\qquad i=L,...,2,\nonumber \\
{\bf H}_{1}\left(x,j_{1};\mathbf{W}\left(k\right)\right) & =\left\langle {\bf w}_{1}\left(k,j_{1}\right),x\right\rangle ,\nonumber 
\end{align}
in which $x\in\mathbb{R}^{d}$ is the input, $\mathbf{W}\left(k\right)=\left\{ {\bf w}_{1}\left(k,\cdot\right),{\bf w}_{i}\left(k,\cdot,\cdot\right):\;i=2,...,L\right\} $
is the weight with ${\bf w}_{1}\left(k,j_{1}\right)\in\mathbb{R}^{d}$,
${\bf w}_{i}\left(k,j_{i-1},j_{i}\right)\in\mathbb{R}$, $\varphi_{i}:\;\mathbb{R}\to\mathbb{R}$
is the activation. Here the network has widths $\left\{ n_{i}\right\} _{i\leq L}$
with $n_{L}=1$, and $k\in\mathbb{N}_{\geq0}$ denotes the time, i.e.
we shall let the network evolve in (discrete) time.

We train the network with stochastic gradient descent (SGD) w.r.t.
the loss ${\cal L}:\;\mathbb{R}\times\mathbb{R}\to\mathbb{R}_{\geq0}$.
We assume that at each time $k$, we draw independently a fresh sample
$z\left(k\right)=\left(x\left(k\right),y\left(k\right)\right)\in\mathbb{R}^{d}\times\mathbb{R}$
from a training distribution ${\cal P}$. Given an initialization
$\mathbf{W}\left(0\right)$, we update $\mathbf{W}\left(k\right)$
according to
\begin{align*}
{\bf w}_{i}\left(k+1,j_{i-1},j_{i}\right) & ={\bf w}_{i}\left(k,j_{i-1},j_{i}\right)-\epsilon\ensuremath{\xi}_{i}\left(t\epsilon\right)\Delta_{i}^{\mathbf{w}}\left(z\left(k\right),j_{i-1},j_{i};\mathbf{W}\left(k\right)\right),\qquad i=2,...,L,\\
{\bf w}_{1}\left(k+1,j_{1}\right) & ={\bf w}_{1}\left(k,j_{1}\right)-\epsilon\ensuremath{\xi}_{1}\left(t\epsilon\right)\Delta_{1}^{\mathbf{w}}\left(z\left(k\right),j_{1};\mathbf{W}\left(k\right)\right),
\end{align*}
in which $j_{i}\in\left[n_{i}\right]$, $\epsilon\in\mathbb{R}_{>0}$
is the learning rate, $\xi_{i}:\;\mathbb{R}_{\geq0}\mapsto\mathbb{R}_{\geq0}$
is the learning rate schedule for $\mathbf{w}_{i}$, and for $z=\left(x,y\right)$,
we define
\begin{align*}
\Delta_{L}^{\mathbf{H}}\left(z,1;\mathbf{W}\left(k\right)\right) & =\partial_{2}{\cal L}\left(y,\hat{\mathbf{y}}\left(x;\mathbf{W}\left(k\right)\right)\right)\varphi_{L}'\left(\mathbf{H}_{L}\left(x,1;\mathbf{W}\left(k\right)\right)\right),\\
\Delta_{i-1}^{\mathbf{H}}\left(z,j_{i-1};\mathbf{W}\left(k\right)\right) & =\frac{1}{n_{i}}\sum_{j_{i}=1}^{n_{i}}\Delta_{i}^{\mathbf{H}}\left(z,j_{i};\mathbf{W}\left(k\right)\right){\bf w}_{i}\left(k,j_{i-1},j_{i}\right)\varphi_{i-1}'\left({\bf H}_{i-1}\left(x,j_{i-1};\mathbf{W}\left(k\right)\right)\right),\qquad i=L,...,2,\\
\Delta_{i}^{\mathbf{w}}\left(z,j_{i-1},j_{i};\mathbf{W}\left(k\right)\right) & =\Delta_{i}^{\mathbf{H}}\left(z,j_{i};\mathbf{W}\left(k\right)\right)\varphi_{i-1}\left({\bf H}_{i-1}\left(x,j_{i-1};\mathbf{W}\left(k\right)\right)\right),\qquad i=L,...,2,\\
\Delta_{1}^{\mathbf{w}}\left(z,j_{1};\mathbf{W}\left(k\right)\right) & =\Delta_{1}^{\mathbf{H}}\left(z,j_{1};\mathbf{W}\left(k\right)\right)x.
\end{align*}
In short, for an initialization $\mathbf{W}\left(0\right)$, we obtain
an SGD trajectory $\mathbf{W}\left(k\right)$ of an $L$-layer network
with size $\left\{ n_{i}\right\} _{i\leq L}$.

\subsection{Mean field limit\label{subsec:Mean-field-limit}}

The MF limit is a continuous-time infinite-width analog of the neural
network under training. We first recall from \cite{nguyen2020rigorous}
the concept of a neuronal ensemble. Given a product probability space
$\left(\Omega,P\right)=\prod_{i=1}^{L}\left(\Omega_{i},P_{i}\right)$
with $\Omega_{L}=\left\{ 1\right\} $, we independently sample $C_{i}\sim P_{i}$,
$i=1,...,L$. In the following, we use $\mathbb{E}_{C_{i}}$ to denote
the expectation w.r.t. the random variable $C_{i}\sim P_{i}$ and
$c_{i}$ to denote a dummy variable $c_{i}\in\Omega_{i}$. The space
$\left(\Omega,P\right)$ is called a neuronal ensemble.

Given a neuronal ensemble $\left(\Omega,P\right)$, the MF limit is
described by a time-evolving system with parameter $W\left(t\right)$,
where the time $t\in\mathbb{R}_{\geq0}$ and $W\left(t\right)=\left\{ w_{1}\left(t,\cdot\right),w_{i}\left(t,\cdot,\cdot\right):\;i=2,...,L\right\} $
with $w_{1}:\,\mathbb{R}_{\geq0}\times\Omega_{1}\to\mathbb{R}^{d}$
and $w_{i}:\,\mathbb{R}_{\geq0}\times\Omega_{i-1}\times\Omega_{i}\to\mathbb{R}$.
It entails the quantities: 
\begin{align}
\hat{y}\left(x;W\left(t\right)\right) & =\varphi_{L}\left(H_{L}\left(x,1;W\left(t\right)\right)\right),\label{eq:three-layer-nn-1}\\
H_{i}\left(x,c_{i};W\left(t\right)\right) & =\mathbb{E}_{C_{i-1}}\left[w_{i}\left(t,C_{i-1},c_{i}\right)\varphi_{i-1}\left(H_{i-1}\left(x,C_{i-1};W\left(t\right)\right)\right)\right],\qquad i=L,...,2,\nonumber \\
H_{1}\left(x,c_{1};W\left(t\right)\right) & =\left\langle w_{1}\left(t,c_{1}\right),x\right\rangle .\nonumber 
\end{align}
The MF limit evolves according to a continuous-time dynamics, described
by a system of ODEs, which we refer to as the MF ODEs. Specifically,
given an initialization $W\left(0\right)=\left\{ w_{1}\left(0,\cdot\right),w_{i}\left(0,\cdot,\cdot\right):\;i=2,...,L\right\} $,
the dynamics solves:
\begin{align*}
\frac{\partial}{\partial t}w_{i}\left(t,c_{i-1},c_{i}\right) & =-\xi_{i}\left(t\right)\mathbb{E}_{Z}\left[\Delta_{i}^{w}\left(Z,c_{i-1},c_{i};W\left(t\right)\right)\right],\qquad i=2,...,L,\\
\frac{\partial}{\partial t}w_{1}\left(t,c_{1}\right) & =-\xi_{1}\left(t\right)\mathbb{E}_{Z}\left[\Delta_{1}^{w}\left(Z,c_{1};W\left(t\right)\right)\right].
\end{align*}
Here $c_{i}\in\Omega_{i}$, $\mathbb{E}_{Z}$ denotes the expectation
w.r.t. the data $Z=\left(X,Y\right)\sim{\cal P}$, and for $z=\left(x,y\right)$,
we define
\begin{align*}
\Delta_{L}^{H}\left(z,1;W\left(t\right)\right) & =\partial_{2}{\cal L}\left(y,\hat{y}\left(x;W\left(t\right)\right)\right)\varphi_{L}'\left(H_{L}\left(x,1;W\left(t\right)\right)\right),\\
\Delta_{i-1}^{H}\left(z,c_{i-1};W\left(t\right)\right) & =\mathbb{E}_{C_{i}}\left[\Delta_{i}^{H}\left(z,C_{i};W\left(t\right)\right)w_{i}\left(t,c_{i-1},C_{i}\right)\varphi_{i-1}'\left(H_{i-1}\left(x,c_{i-1};W\left(t\right)\right)\right)\right],\qquad i=L,...,2,\\
\Delta_{i}^{w}\left(z,c_{i-1},c_{i};W\left(t\right)\right) & =\Delta_{i}^{H}\left(z,c_{i};W\left(t\right)\right)\varphi_{i-1}\left(H_{i-1}\left(x,c_{i-1};W\left(t\right)\right)\right),\qquad i=L,...,2,\\
\Delta_{1}^{w}\left(z,c_{1};W\left(t\right)\right) & =\Delta_{1}^{H}\left(z,c_{1};W\left(t\right)\right)x.
\end{align*}
In short, given a neuronal ensemble $\left(\Omega,P\right)$, for
each initialization $W\left(0\right)$, we have defined a MF limit
$W\left(t\right)$.

\section{Convergence to global optima\label{sec:global_conv}}

\subsection{Main result: global convergence}

To measure the learning quality, we consider the loss averaged over
the data $Z\sim{\cal P}$:
\[
\mathscr{L}\left(F\right)=\mathbb{E}_{Z}\left[{\cal L}\left(Y,\hat{y}\left(X;F\right)\right)\right],
\]
where $F=\left\{ f_{i}:\;i=1,...,L\right\} $ a set of measurable
functions $f_{1}:\;\Omega_{1}\to\mathbb{R}^{d}$, $f_{i}:\;\Omega_{i-1}\times\Omega_{i}\to\mathbb{R}$
for $i=2,...,L$.

We also recall the concept of a neuronal embedding from \cite{nguyen2020rigorous}.
Formally, in the present context, it is a tuple $\left(\Omega,P,\left\{ w_{i}^{0}\right\} _{i\leq L}\right)$,
comprising of a neuronal ensemble $\left(\Omega,P\right)$ and a set
of measurable functions $\left\{ w_{i}^{0}\right\} _{i\leq L}$ in
which $w_{1}^{0}:\;\Omega_{1}\to\mathbb{R}^{d}$ and $w_{i}^{0}:\;\Omega_{i-1}\times\Omega_{i}\to\mathbb{R}$
for $i=2,...,L$. The neuronal embedding connects a finite-width neural
network and its MF limit, via their initializations which are specified
by $\left\{ w_{i}^{0}\right\} _{i\leq L}$. We shall revisit this
connection later in Section \ref{sec:connection}. In the following,
we focus on the analysis of the MF limit.
\begin{assumption}
\label{assump:multilayer}Consider a neuronal embedding $\left(\Omega,P,\left\{ w_{i}^{0}\right\} _{i\leq L}\right)$,
recalling $\Omega=\prod_{i=1}^{L}\Omega_{i}$ and $P=\prod_{i=1}^{L}P_{i}$
with $\Omega_{L}=\left\{ 1\right\} $. Consider the MF limit associated
with the neuronal ensemble $\left(\Omega,P\right)$ with initialization
$W\left(0\right)$ such that $w_{1}\left(0,\cdot\right)=w_{1}^{0}\left(\cdot\right)$
and $w_{i}\left(0,\cdot,\cdot\right)=w_{i}^{0}\left(\cdot,\cdot\right)$.
We make the following assumptions:
\begin{enumerate}
\item Regularity: We assume that
\begin{itemize}
\item $\varphi_{i}$ is $K$-bounded for $1\leq i\leq L-1$, $\varphi_{i}'$
is $K$-bounded and $K$-Lipschitz for $1\leq i\leq L$, and $\varphi_{L}'$
is non-zero everywhere,
\item $\partial_{2}{\cal L}\left(\cdot,\cdot\right)$ is $K$-Lipschitz
in the second variable and $K$-bounded,
\item $\left|X\right|\leq K$ with probability $1$,
\item $\xi_{i}$ is $K$-bounded and $K$-Lipschitz for $1\leq i\leq L$,
\item $\sup_{k\geq1}k^{-1/2}\mathbb{E}\left[\left|w_{1}^{0}\left(C_{1}\right)\right|^{k}\right]^{1/k}\leq K$
and $\sup_{k\geq1}k^{-1/2}\mathbb{E}\left[\left|w_{i}^{0}\left(C_{i-1},C_{i}\right)\right|^{k}\right]^{1/k}\leq K$
for $i=2,...,L$.
\end{itemize}
\item Diversity: The functions $\left\{ w_{i}^{0}\right\} _{i\leq L}$ satisfy
that
\begin{itemize}
\item ${\rm supp}\left(w_{1}^{0}\left(C_{1}\right),w_{2}^{0}\left(C_{1},\cdot\right)\right)=\mathbb{R}^{d}\times L^{2}\left(P_{2}\right)$,
\item ${\rm supp}\left(w_{i}^{0}\left(\cdot,C_{i}\right),w_{i+1}^{0}\left(C_{i},\cdot\right)\right)=L^{2}\left(P_{i-1}\right)\times L^{2}\left(P_{i+1}\right)$
for $i=2,...,L-1$.
\end{itemize}
(Remark: we write $w_{i}^{0}\left(\cdot,C_{i}\right)$ to denote the
random mapping $c_{i-1}\mapsto w_{i}^{0}\left(c_{i-1},C_{i}\right)$,
and similar for $w_{i+1}^{0}\left(C_{i},\cdot\right)$.)
\item Convergence: There exist limits $\left\{ \bar{w}_{i}\right\} _{i\leq L}$
such that as $t\to\infty$,
\begin{align*}
\mathbb{E}\left[\left|w_{i}\left(t,C_{i-1},C_{i}\right)-\bar{w}_{i}\left(C_{i-1},C_{i}\right)\right|\sideset{}{_{j=i+1}^{L}}\prod\left|\bar{w}_{j}\left(C_{j-1},C_{j}\right)\right|\right] & \to0,\quad i=2,...,L,\\
\mathbb{E}\left[\left|w_{1}\left(t,C_{1}\right)-\bar{w}_{1}\left(C_{1}\right)\right|\sideset{}{_{j=2}^{L}}\prod\left|\bar{w}_{j}\left(C_{j-1},C_{j}\right)\right|\right] & \to0,\\
{\rm ess\text{-}sup}\left|\frac{\partial}{\partial t}w_{L}\left(t,C_{L-1},1\right)\right| & \to0.
\end{align*}
(Here we take $\prod_{j=i+1}^{L}=1$ for $i=L$.)
\item Universal approximation: The set $\left\{ \varphi_{1}\left(\left\langle u,\cdot\right\rangle \right):\;u\in\mathbb{R}^{d}\right\} $
has dense span in $L^{2}\left({\cal P}_{X}\right)$ (the space of
square integrable functions w.r.t. the measure ${\cal P}_{X}$, which
is the distribution of the input $X$). Furthermore, for each $i=2,...,L-1$,
$\varphi_{i}$ is non-obstructive in the sense that the set $\left\{ \varphi_{i}\circ f:\;f\in L^{2}\left({\cal P}_{X}\right)\right\} $
has dense span in $L^{2}\left({\cal P}_{X}\right)$.
\end{enumerate}
\end{assumption}

The first assumption can be satisfied for several common setups and
loss functions. The third assumption, similar to \cite{chizat2018,nguyen2020rigorous},
is technical and sets the focus on settings where the MF dynamics
converges with time, although we note it is an assumption on the mode
of convergence only and not on the limits $\left\{ \bar{w}_{i}\right\} _{i\leq L}$.
The fourth assumption is natural and can be satisfied by common activations.
For example, $\varphi_{i}$ can be $\tanh$ for $i=1,...,L-1$. In
general, for a bounded and continuous $\varphi_{i}$ to be non-obstructive,
it suffices that $\varphi_{i}$ is not a constant function. The second
assumption is new: it refers to an initialization scheme that introduces
correlation among the weights. In particular, i.i.d. initializations
do not satisfy this assumption for $L\geq3$.
\begin{thm}
\label{thm:existence}Given any neuronal ensemble $\left(\Omega,P\right)$
and a set of functions $\left\{ w_{i}^{0}\right\} _{i\leq L}$ such
that the regularity assumption listed in Assumption \ref{assump:multilayer}
is satisfied, and given an initialization $W\left(0\right)$ such
that $w_{1}\left(0,\cdot\right)=w_{1}^{0}\left(\cdot\right)$ and
$w_{i}\left(0,\cdot,\cdot\right)=w_{i}^{0}\left(\cdot,\cdot\right)$,
there exists a unique solution $W$ to the MF ODEs on $t\in[0,\infty)$.
\end{thm}

This theorem can be proven in a similar manner to \cite[Theorem 3]{nguyen2020rigorous},
so we will not show the complete proof here. The main focus is on
the global convergence result, which we state next.

\begin{thm}
\label{thm:global-optimum}Consider a neuronal embedding $\left(\Omega,P,\left\{ w_{i}^{0}\right\} _{i\leq L}\right)$
and the MF limit as in Assumption \ref{assump:multilayer}. Assume
$\xi_{L}\left(\cdot\right)=1$. Then:
\begin{itemize}
\item Case 1 (convex loss): If ${\cal L}$ is convex in the second variable,
then $\left\{ \bar{w}_{i}\right\} _{i\leq L}$ is a global minimizer
of $\mathscr{L}$:
\[
\mathscr{L}\left(\left\{ \bar{w}_{i}\right\} _{i\leq L}\right)=\inf_{F}\mathscr{L}\left(F\right)=\inf_{\tilde{y}:\;\mathbb{R}^{d}\to\mathbb{R}}\mathbb{E}_{Z}\left[{\cal L}\left(Y,\tilde{y}\left(X\right)\right)\right].
\]
\item Case 2 (generic non-negative loss): Suppose that $\partial_{2}{\cal L}\left(y,\hat{y}\right)=0$
implies ${\cal L}\left(y,\hat{y}\right)=0$. If $y=y(x)$ is a function
of $x$, then $\mathscr{L}\left(\left\{ \bar{w}_{i}\right\} _{i\leq L}\right)=0$.
\end{itemize}
\end{thm}

The assumptions here are similar to those made in \cite{nguyen2020rigorous}.
We remark on a special difference. In \cite{nguyen2020rigorous},
the diversity assumption refers to a full support condition of the
first layer's weight only. Here our diversity assumptions refers to
a certain full support condition for all layers. At a closer look,
the condition is in the function space and reflects certain \textsl{bidirectional
diversity}. In particular, this assumption implies both $w_{i}^{0}\left(\cdot,C_{i}\right)$
and $w_{i}^{0}\left(C_{i-1},\cdot\right)$ have full supports in $L^{2}\left(P_{i-1}\right)$
and $L^{2}\left(P_{i}\right)$ respectively (which we shall refer
to as \textit{forward diversity} and \textit{backward diversity},
respectively), for $2\leq i\leq L-1$.

The proof proceeds with several insights that have already appeared
in \cite{nguyen2020rigorous}. The novelty of our present analysis
lies in the use of the aforementioned bidirectional diversity. To
clarify the point, let us give a brief high-level idea of the proof.
At time $t$ sufficiently large, we expect to have:
\[
\left|\frac{\partial}{\partial t}w_{L}\left(t,c_{L-1},1\right)\right|=\left|\mathbb{E}_{Z}\left[\partial_{2}{\cal L}\left(Y,\hat{y}\left(X;W\left(t\right)\right)\right)\varphi_{L}'\left(H_{L}\left(X,1;W\left(t\right)\right)\right)\varphi_{L-1}\left(H_{L-1}\left(X,c_{L-1};W\left(t\right)\right)\right)\right]\right|\approx0
\]
for $P_{L-1}$-almost every $c_{L-1}$. If the set of mappings $x\mapsto H_{L-1}\left(x,c_{L-1};W\left(t\right)\right)$,
indexed by $c_{L-1}$, is diverse in the sense that ${\rm supp}\left(H_{L-1}\left(\cdot,C_{L-1};W\left(t\right)\right)\right)=L^{2}\left({\cal P}_{X}\right)$,
then since $\varphi_{L-1}$ is non-obstructive, we obtain
\[
\mathbb{E}_{Z}\left[\partial_{2}{\cal L}\left(Y,\hat{y}\left(X;W\left(t\right)\right)\right)\middle|X=x\right]\varphi_{L}'\left(H_{L}\left(x,1;W\left(t\right)\right)\right)\approx0
\]
and consequently
\[
\mathbb{E}_{Z}\left[\partial_{2}{\cal L}\left(Y,\hat{y}\left(X;W\left(t\right)\right)\right)\middle|X=x\right]\approx0
\]
for ${\cal P}_{X}$-almost every $x$. The desired conclusion then
follows.

Hence the crux of the proof is to show that ${\rm supp}\left(H_{L-1}\left(\cdot,C_{L-1};W\left(t\right)\right)\right)=L^{2}\left({\cal P}_{X}\right)$.
In fact, we show that this holds for any finite time $t\geq0$. This
follows if we can prove the forward diversity property of the weights,
in which $w_{i}\left(t,\cdot,C_{i}\right)$ has full support in $L^{2}\left(P_{i-1}\right)$
for any $t\geq0$ and $2\leq i\leq L-1$, and a similar property for
$w_{1}\left(t,C_{1}\right)$. Interestingly to that end, we actually
show that bidirectional diversity, and hence both forward diversity
and backward diversity, hold at any time $t\geq0$, even though we
only need forward diversity for our purpose.

\subsection{Proof of Theorem \ref{thm:global-optimum}}
\begin{proof}
We divide the proof into several steps.

\paragraph*{Step 1: Diversity of the weights.}

We show that ${\rm supp}\left(w_{1}\left(t,C_{1}\right)\right)=\mathbb{R}^{d}$
and ${\rm supp}\left(w_{i}\left(t,\cdot,C_{i}\right)\right)=L^{2}\left(P_{i-1}\right)$
for $i=2,...,L-1$, for any $t\geq0$. We do so by showing a stronger
statement, that the following bidirectional diversity condition holds
at any finite training time: 
\begin{align*}
{\rm supp}\left(w_{1}\left(t,C_{1}\right),w_{2}\left(t,C_{1},\cdot\right)\right) & =\mathbb{R}^{d}\times L^{2}\left(P_{2}\right),\\
{\rm supp}\left(w_{i}\left(t,\cdot,C_{i}\right),w_{i+1}\left(t,C_{i},\cdot\right)\right) & =L^{2}\left(P_{i-1}\right)\times L^{2}\left(P_{i+1}\right),\qquad i=2,...,L-1,
\end{align*}
for any $t\geq0$.

We prove the first statement. Given a MF trajectory $\left(W\left(t\right)\right)_{t\ge0}$
and $u_{1}\in\mathbb{R}^{d}$, $u_{2}\in L^{2}\left(P_{2}\right)$,
we consider the following flow on $\mathbb{R}^{d}\times L^{2}\left(P_{2}\right)$:
\begin{align}
\frac{\partial}{\partial t}a_{2}^{+}\left(t,c_{2};u\right) & =-\xi_{2}(t)\mathbb{E}_{Z}\left[\Delta_{2}^{H}\left(Z,c_{2};W(t)\right)\varphi_{1}\left(\left\langle a_{1}^{+}\left(t;u\right),X\right\rangle \right)\right],\nonumber \\
\frac{\partial}{\partial t}a_{1}^{+}\left(t;u\right) & =-\xi_{1}(t)\mathbb{E}_{Z}\left[\mathbb{E}_{C_{2}}\left[\Delta_{2}^{H}\left(Z,C_{2};W(t)\right)a_{2}^{+}\left(t,C_{2};u\right)\right]\varphi_{1}'\left(\left\langle a_{1}^{+}\left(t;u\right),X\right\rangle \right)X\right],\label{eq:ODE-w}
\end{align}
for $u=\left(u_{1},u_{2}\right)$, with the initialization $a_{1}^{+}\left(0;u\right)=u_{1}$
and $a_{2}^{+}\left(0,c_{2};u\right)=u_{2}\left(c_{2}\right)$. Existence
and uniqueness of $\left(a_{1}^{+},a_{2}^{+}\right)$ follows similarly
to Theorem \ref{thm:existence}. We next prove for all finite $T>0$
and $u^{+}=\left(u_{1}^{+},u_{2}^{+}\right)\in\mathbb{R}^{d}\times L^{2}\left(P_{2}\right)$,
there exists $u^{-}=\left(u_{1}^{-},u_{2}^{-}\right)\in\mathbb{R}^{d}\times L^{2}\left(P_{2}\right)$
such that 
\[
a_{1}^{+}\left(T;u^{-}\right)=u_{1}^{+},\qquad a_{2}^{+}\left(T,\cdot;u^{-}\right)=u_{2}^{+}.
\]
We consider the following auxiliary dynamics on $\mathbb{R}^{d}\times L^{2}\left(P_{2}\right)$:
\begin{align}
\frac{\partial}{\partial t}a_{2}^{-}\left(t,c_{2};u\right) & =\xi_{2}(T-t)\mathbb{E}_{Z}\left[\Delta_{2}^{H}\left(Z,c_{2};W(T-t)\right)\varphi_{1}\left(\left\langle a_{1}^{-}\left(t;u\right),X\right\rangle \right)\right],\nonumber \\
\frac{\partial}{\partial t}a_{1}^{-}\left(t;u\right) & =\xi_{1}(T-t)\mathbb{E}_{Z}\left[\mathbb{E}_{C_{2}}\left[\Delta_{2}^{H}\left(Z,C_{2};W(T-t)\right)a_{2}^{-}\left(t,C_{2};u\right)\right]\varphi_{1}'\left(\left\langle a_{1}^{-}\left(t;u\right),X\right\rangle \right)X\right],\label{eq:ODE-a}
\end{align}
initialized at $a_{1}^{-}\left(0;u\right)=u_{1}$ and $a_{2}^{-}\left(0,c_{2};u\right)=u_{2}\left(c_{2}\right)$,
for $u=\left(u_{1},u_{2}\right)\in\mathbb{R}^{d}\times L^{2}\left(P_{2}\right)$.
Existence and uniqueness of $(a_{1}^{-},a_{2}^{-})$ follow similarly
to Theorem \ref{thm:existence}. Observe that the pair
\[
\tilde{a}_{1}^{-}\left(t\right)=a_{1}^{-}\left(T-t;u^{+}\right),\qquad\tilde{a}_{2}^{-}\left(t,c_{2}\right)=a_{2}^{-}\left(T-t,c_{2};u^{+}\right)
\]
solves the system 
\begin{align*}
\frac{\partial}{\partial t}\tilde{a}_{2}^{-}\left(t,c_{2}\right) & =-\frac{\partial}{\partial t}a_{2}^{-}\left(T-t,c_{2};u^{+}\right)=-\xi_{2}(t)\mathbb{E}_{Z}\left[\Delta_{2}^{H}\left(Z,c_{2};W(t)\right)\varphi_{1}\left(\left\langle \tilde{a}_{1}^{-}\left(t\right),X\right\rangle \right)\right],\\
\frac{\partial}{\partial t}\tilde{a}_{1}^{-}\left(t\right) & =-\frac{\partial}{\partial t}a_{1}^{-}\left(T-t;u^{+}\right)=-\xi_{1}(t)\mathbb{E}_{Z}\left[\mathbb{E}_{C_{2}}\left[\Delta_{2}^{H}\left(Z,C_{2};W(t)\right)\tilde{a}_{2}^{-}\left(t,C_{2}\right)\right]\varphi_{1}'\left(\left\langle \tilde{a}_{1}^{-}\left(t\right),X\right\rangle \right)X\right],
\end{align*}
initialized at $\tilde{a}_{2}^{-}\left(0,c_{2}\right)=a_{2}^{-}\left(T,c_{2};u^{+}\right)$
and $\tilde{a}_{1}^{-}(0)=a_{1}^{-}\left(T;u^{+}\right)$. Thus, by
uniqueness of the solution to the ODE (\ref{eq:ODE-w}), $(\tilde{a}_{1}^{-},\tilde{a}_{2}^{-})$
forms a solution of the ODE (\ref{eq:ODE-w}) initialized at 
\[
\tilde{a}_{1}^{-}(0)=a_{1}^{-}\left(T;u^{+}\right),\qquad\tilde{a}_{2}^{-}\left(0,c_{2}\right)=a_{2}^{-}\left(T,c_{2};u^{+}\right).
\]
In particular, the solution $(\tilde{a}_{1}^{-},\tilde{a}_{2}^{-})$
of the ODE (\ref{eq:ODE-w}) with this initialization satisfies 
\[
\tilde{a}_{1}^{-}\left(T\right)=a_{1}^{-}\left(0;u^{+}\right)=u_{1}^{+},\qquad\tilde{a}_{2}^{-}\left(T,\cdot\right)=a_{2}^{-}\left(0,\cdot;u^{+}\right)=u_{2}^{+}.
\]
Let $u_{1}^{-}=a_{1}^{-}\left(T;u^{+}\right)$ and $u_{2}^{-}=a_{2}^{-}\left(T,\cdot;u^{+}\right)$.
Then we have $a_{1}^{+}\left(T;u^{-}\right)=u_{1}^{+}$ and $a_{2}^{+}\left(T,\cdot;u^{-}\right)=u_{2}^{+}$
as desired.

Using this, by continuity of the map $u\mapsto\left(a_{1}^{+}\left(T;u\right),a_{2}^{+}\left(T,\cdot;u\right)\right)$,
for every $\epsilon>0$, there exists a neighborhood $U$ of $u^{-}$
such that for any $u\in U$, $\left|\left(a_{1}^{+}\left(T;u\right),a_{2}^{+}\left(T,\cdot;u\right)\right)-u^{+}\right|\le\epsilon$.
Notice that the MF trajectory $W\left(t\right)$ satisfies 
\[
w_{1}\left(t,c_{1}\right)=a_{1}^{+}\left(t;w_{1}\left(0,c_{1}\right),w_{2}\left(0,c_{1},\cdot\right)\right),\qquad w_{2}\left(t,c_{1},\cdot\right)=a_{2}^{+}\left(t,\cdot;w_{1}\left(0,c_{1}\right),w_{2}\left(0,c_{1},\cdot\right)\right).
\]
Then since $\left(w_{1}\left(0,C_{1}\right),w_{2}\left(0,C_{1},\cdot\right)\right)$
has full support in $\mathbb{R}^{d}\times L^{2}\left(P_{2}\right)$,
for any finite $T>0$, we have $\left(w_{1}\left(T,C_{1}\right),w_{2}\left(T,C_{1},\cdot\right)\right)$
has full support in $\mathbb{R}^{d}\times L^{2}\left(P_{2}\right)$,
proving the first statement.

The other statements can be proven similarly by considering the following
pairs of flows on $L^{2}(P_{i-1})\times L^{2}(P_{i+1})$, for $u=\left(u_{1},u_{2}\right)\in L^{2}(P_{i-1})\times L^{2}(P_{i+1})$:
\begin{align*}
\frac{\partial}{\partial t}a_{i}^{+}\left(t,c_{i-1};u\right) & =-\xi_{i}\left(t\right)\mathbb{E}_{Z}\left[\Delta_{i}^{a}\left(Z,a_{i}^{+}\left(t,\cdot;u\right),a_{i+1}^{+}\left(t,\cdot;u\right);W\left(t\right)\right)\varphi_{i-1}\left(H_{i-1}\left(X,c_{i-1};W\left(t\right)\right)\right)\right],\\
\frac{\partial}{\partial t}a_{i+1}^{+}\left(t,c_{i+1};u\right) & =-\xi_{i+1}\left(t\right)\mathbb{E}_{Z}\left[\Delta_{i+1}^{H}\left(Z,c_{i+1};W\left(t\right)\right)\varphi_{i}\left(H_{i}^{a}\left(Z,a_{i}^{+}\left(t,\cdot;u\right);W\left(t\right)\right)\right)\right],
\end{align*}
initialized at $a_{i}^{+}\left(0,c_{i-1};u\right)=u_{1}\left(c_{i-1}\right)$
and $a_{i+1}^{+}\left(0,c_{i+1};u\right)=u_{2}\left(c_{i+1}\right)$,
and
\begin{align*}
\frac{\partial}{\partial t}a_{i}^{-}\left(t,c_{i-1};u\right) & =\xi_{i}\left(T-t\right)\mathbb{E}_{Z}\left[\Delta_{i}^{a}\left(Z,a_{i}^{-}\left(t,\cdot;u\right),a_{i+1}^{-}\left(t,\cdot;u\right);W\left(T-t\right)\right)\varphi_{i-1}\left(H_{i-1}\left(X,c_{i-1};W\left(T-t\right)\right)\right)\right],\\
\frac{\partial}{\partial t}a_{i+1}^{-}\left(t,c_{i+1};u\right) & =\xi_{i+1}\left(T-t\right)\mathbb{E}_{Z}\left[\Delta_{i+1}^{H}\left(Z,c_{i+1};W(T-t)\right)\varphi_{i}\left(H_{i}^{a}\left(Z,a_{i}^{-}\left(t,\cdot;u\right);W\left(T-t\right)\right)\right)\right],
\end{align*}
initialized at $a_{i}^{-}\left(0,c_{i-1};u\right)=u_{1}\left(c_{i-1}\right)$
and $a_{i+1}^{-}\left(0,c_{i+1};u\right)=u_{2}\left(c_{i+1}\right)$,
in which we define:
\begin{align*}
\Delta_{i}^{a}\left(z,f,g;W\left(t\right)\right) & =\mathbb{E}_{C_{i+1}}\left[\Delta_{i+1}^{H}\left(z,C_{i+1};W\left(t\right)\right)g\left(C_{i+1}\right)\varphi_{i}'\left(H_{i}^{a}\left(z,f;W\left(t\right)\right)\right)\right],\\
H_{i}^{a}\left(z,f;W\left(t\right)\right) & =\mathbb{E}_{C_{i-1}}\left[f\left(C_{i-1}\right)\varphi_{i-1}\left(H_{i-1}\left(x,C_{i-1};W\left(t\right)\right)\right)\right],
\end{align*}
for $f\in L^{2}\left(P_{i-1}\right)$ and $g\in L^{2}\left(P_{i+1}\right)$.

\paragraph*{Step 2: Diversity of the pre-activations.}

We show that ${\rm supp}\left(H_{i}\left(\cdot,C_{i};W\left(t\right)\right)\right)=L^{2}\left({\cal P}_{X}\right)$
for any $t\geq0$, for $i=2,...,L-1$ by induction.

Firstly consider the base case $i=2$. Recall that
\[
H_{2}\left(x,c_{2};W\left(t\right)\right)=\mathbb{E}_{C_{1}}\left[w_{2}\left(t,C_{1},c_{2}\right)\varphi_{1}\left(\left\langle w_{1}\left(t,C_{1}\right),x\right\rangle \right)\right]\equiv{\cal H}_{2}\left(t,x,w_{2}\left(t,\cdot,c_{2}\right)\right).
\]
Observe that the set ${\rm cl}\left(\left\{ {\cal H}_{2}\left(t,\cdot,f\right):\;f\in L^{2}\left(P_{1}\right)\right\} \right)$
is a closed linear subspace of $L^{2}\left({\cal P}_{X}\right)$.
Hence this set is equal to $L^{2}\left({\cal P}_{X}\right)$ if it
has dense span in $L^{2}\left({\cal P}_{X}\right)$, which we show
now. Indeed, suppose that for some $g\in L^{2}\left({\cal P}_{X}\right)$
such that $\left|g\right|\neq0$, we have $\mathbb{E}_{Z}\left[g\left(X\right){\cal H}_{2}\left(t,X,f\right)\right]=0$
for all $f\in L^{2}\left(P_{1}\right)$. Equivalently,
\[
\mathbb{E}_{C_{1}}\left[f\left(C_{1}\right)\mathbb{E}_{Z}\left[g\left(X\right)\varphi_{1}\left(\left\langle w_{1}\left(t,C_{1}\right),X\right\rangle \right)\right]\right]=0,
\]
for all $f\in L^{2}\left(P_{1}\right)$. As such, for $P_{1}$-almost
every $c_{1}$,
\[
\mathbb{E}_{Z}\left[g\left(X\right)\varphi_{1}\left(\left\langle w_{1}\left(t,c_{1}\right),X\right\rangle \right)\right]=0.
\]
Since ${\rm supp}\left(w_{1}\left(t,C_{1}\right)\right)=\mathbb{R}^{d}$
and that the mapping $u\mapsto\varphi_{1}\left(\left\langle u,x\right\rangle \right)$
is continuous, by the universal approximation assumption for $\varphi_{1}$,
we then obtain $g(x)=0$ for $P_{X}$-almost every $x$, which is
a contradiction. We have thus proved that ${\rm cl}\left(\left\{ {\cal H}_{2}\left(t,\cdot,f\right):\;f\in L^{2}\left(P_{1}\right)\right\} \right)=L^{2}\left({\cal P}_{X}\right)$.
Note that $f\mapsto{\cal H}_{2}\left(t,x,f\right)$ is continuous,
and ${\rm supp}\left(w_{2}\left(t,\cdot,C_{2}\right)\right)=L^{2}\left(P_{1}\right)$,
we then have ${\rm supp}\left(H_{2}\left(\cdot,C_{2};W\left(t\right)\right)\right)=L^{2}\left({\cal P}_{X}\right)$
as desired.

Now let us assume that ${\rm supp}\left(H_{i-1}\left(\cdot,C_{i-1};W\left(t\right)\right)\right)=L^{2}\left({\cal P}_{X}\right)$
for some $i\geq3$ (the induction hypothesis). We would like to show
${\rm supp}\left(H_{i}\left(\cdot,C_{i};W\left(t\right)\right)\right)=L^{2}\left({\cal P}_{X}\right)$.
This is similar to the base case. In particular, recall that
\[
H_{i}\left(x,c_{i};W\left(t\right)\right)=\mathbb{E}_{C_{i-1}}\left[w_{i}\left(t,C_{i-1},c_{i}\right)\varphi_{i-1}\left(H_{i-1}\left(x,C_{i-1};W\left(t\right)\right)\right)\right]\equiv{\cal H}_{i}\left(t,x,w_{i}\left(t,\cdot,c_{i}\right)\right).
\]
Now suppose that for some $g\in L^{2}\left({\cal P}_{X}\right)$ such
that $\left|g\right|\neq0$, we have $\mathbb{E}_{Z}\left[g\left(X\right){\cal H}_{i}\left(t,X,f\right)\right]=0$
for all $f\in L^{2}\left(P_{i-1}\right)$. Then, for $P_{i-1}$-almost
every $c_{i-1}$,
\[
\mathbb{E}_{Z}\left[g\left(X\right)\varphi_{i-1}\left(H_{i-1}\left(X,c_{i-1};W\left(t\right)\right)\right)\right]=0.
\]
Recall the induction hypothesis ${\rm supp}\left(H_{i-1}\left(\cdot,C_{i-1};W\left(t\right)\right)\right)=L^{2}\left({\cal P}_{X}\right)$.
Since $\varphi_{i-1}$ is non-obstructive and continuous, we obtain
$g(x)=0$ for $P_{X}$-almost every $x$, which is a contradiction.
Therefore the set ${\rm cl}\left(\left\{ {\cal H}_{i}\left(t,\cdot,f\right):\;f\in L^{2}\left(P_{i-1}\right)\right\} \right)$
has dense span in $L^{2}\left({\cal P}_{X}\right)$, and again, this
implies it is equal to $L^{2}\left({\cal P}_{X}\right)$. Since $f\mapsto{\cal H}_{i}\left(t,x,f\right)$
is continuous and ${\rm supp}\left(w_{i}\left(t,\cdot,C_{i}\right)\right)=L^{2}\left(P_{i-1}\right)$,
we have ${\rm supp}\left(H_{i}\left(\cdot,C_{i};W\left(t\right)\right)\right)=L^{2}\left({\cal P}_{X}\right)$.

\paragraph*{Step 3: Concluding.}

Let $\mathbb{E}_{Z}\left[\partial_{2}{\cal L}\left(Y,\hat{y}\left(X;W\left(t\right)\right)\right)\middle|X=x\right]\varphi_{L}'\left(H_{L}\left(x,1;W\left(t\right)\right)\right)={\cal H}\left(x,W\left(t\right)\right)$.
From the last step, we have ${\rm supp}\left(H_{L-1}\left(\cdot,C_{L-1};W\left(t\right)\right)\right)=L^{2}\left({\cal P}_{X}\right)$
for any $t\geq0$. Recall that
\[
\frac{\partial}{\partial t}w_{L}\left(t,c_{L-1},1\right)=-\mathbb{E}_{Z}\left[{\cal H}\left(X,W\left(t\right)\right)\varphi_{L-1}\left(H_{L-1}\left(X,c_{L-1};W\left(t\right)\right)\right)\right].
\]
By the convergence assumption, for any $\epsilon>0$, there exists
$T\left(\epsilon\right)>0$ such that for any $t\geq T\left(\epsilon\right)$,
for $P_{L-1}$-almost every $c_{L-1}$,
\[
\left|\mathbb{E}_{Z}\left[{\cal H}\left(X,W\left(t\right)\right)\varphi_{L-1}\left(H_{L-1}\left(X,c_{L-1};W\left(t\right)\right)\right)\right]\right|\leq\epsilon.
\]
We claim that ${\cal H}\left(x,W\left(t\right)\right)\to{\cal H}\left(x,\left\{ \bar{w}_{i}\right\} _{i\leq L}\right)$
in $L^{1}\left({\cal P}_{X}\right)$ as $t\to\infty$. Assuming this
claim and recalling that $\varphi_{L-1}$ is $K$-bounded by the regularity
assumption, we then have that for some $T'\left(\epsilon\right)\geq T\left(\epsilon\right)$,
for any $t\geq T'\left(\epsilon\right)$,
\begin{align*}
 & {\rm ess\text{-}sup}\left|\mathbb{E}_{Z}\left[{\cal H}\left(X,\left\{ \bar{w}_{i}\right\} _{i\leq L}\right)\varphi_{L-1}\left(H_{L-1}\left(X,C_{L-1};W\left(t\right)\right)\right)\right]\right|\\
 & \leq K\mathbb{E}_{Z}\left[\left|{\cal H}\left(X,\left\{ \bar{w}_{i}\right\} _{i\leq L}\right)-{\cal H}\left(X,W\left(t\right)\right)\right|\right]+{\rm ess\text{-}sup}\left|\mathbb{E}_{Z}\left[{\cal H}\left(X,W\left(t\right)\right)\varphi_{L-1}\left(H_{L-1}\left(X,C_{L-1};W\left(t\right)\right)\right)\right]\right|\\
 & \leq K\epsilon.
\end{align*}
Since ${\rm supp}\left(H_{L-1}\left(\cdot,C_{L-1};W\left(t\right)\right)\right)=L^{2}\left({\cal P}_{X}\right)$
and $\varphi_{L-1}$ is continuous,
\[
\left|\mathbb{E}_{Z}\left[{\cal H}\left(X,\left\{ \bar{w}_{i}\right\} _{i\leq L}\right)f\left(X\right)\right]\right|\leq K\epsilon\qquad\forall f\in S,
\]
for $S=\left\{ \varphi_{L-1}\circ g:\;g\in L^{2}\left({\cal P}_{X}\right)\right\} $.
Since $\epsilon>0$ is arbitrary,
\[
\left|\mathbb{E}_{Z}\left[{\cal H}\left(X,\left\{ \bar{w}_{i}\right\} _{i\leq L}\right)f\left(X\right)\right]\right|=0\qquad\forall f\in S.
\]
Furthermore, since $\varphi_{L-1}$ is non-obstructive, $S$ has dense
span in $L^{2}\left({\cal P}_{X}\right)$. Therefore ${\cal H}\left(x,\left\{ \bar{w}_{i}\right\} _{i\leq L}\right)=0$
for ${\cal P}_{X}$-almost every $x$. Since $\varphi_{L}'$ is non-zero
everywhere, 
\[
\mathbb{E}_{Z}\left[\partial_{2}{\cal L}\left(Y,\hat{y}\left(X;\left\{ \bar{w}_{i}\right\} _{i\leq L}\right)\right)\middle|X=x\right]=0
\]
 for ${\cal P}_{X}$-almost every $x$.

In Case 1, due to convexity of ${\cal L}$, for any measurable function
$\tilde{y}$:
\[
{\cal L}\left(y,\tilde{y}\left(x\right)\right)-{\cal L}\left(y,\hat{y}\left(x;\left\{ \bar{w}_{i}\right\} _{i\leq L}\right)\right)\geq\partial_{2}{\cal L}\left(y,\hat{y}\left(x,\left\{ \bar{w}_{i}\right\} _{i\leq L}\right)\right)\left(\tilde{y}\left(x\right)-\hat{y}\left(x,\left\{ \bar{w}_{i}\right\} _{i\leq L}\right)\right).
\]
Taking expectation, we get $\mathbb{E}_{Z}\left[{\cal L}\left(Y,\tilde{y}\left(X\right)\right)\right]\geq\mathscr{L}\left(\left\{ \bar{w}_{i}\right\} _{i\leq L}\right)$.

In Case 2, we have $\partial_{2}{\cal L}\left(y\left(x\right),\hat{y}\left(x;\left\{ \bar{w}_{i}\right\} _{i\leq L}\right)\right)=0$,
and hence ${\cal L}\left(y\left(x\right),\hat{y}\left(x;\left\{ \bar{w}_{i}\right\} _{i\leq L}\right)\right)=0$,
for ${\cal P}_{X}$-almost every $x$, since $y$ is a function of
$x$.

We are left with proving the claim that ${\cal H}(x,W(t))\to{\cal H}\left(x,\{\bar{w}_{i}\}_{i\le L}\right)$
in $L^{1}({\cal P}_{X})$ as $t\to\infty$. For brevity, we denote
\[
\delta_{i}\left(t,x,c_{i}\right)=\left|H_{i}\left(x,c_{i};W\left(t\right)\right)-H_{i}\left(x,c_{i};\left\{ \bar{w}_{i}\right\} _{i\leq L}\right)\right|.
\]
First observe that by the regularity assumption, for $2\leq i\leq L$:
\begin{align*}
\delta_{i}\left(t,x,c_{i}\right) & \leq K\mathbb{E}_{C_{i-1}}\left[\left|w_{i}\left(t,C_{i-1},c_{i}\right)-\bar{w}_{i}\left(C_{i-1},c_{i}\right)\right|+\left|\bar{w}_{i}\left(C_{i-1},c_{i}\right)\right|\delta_{i-1}\left(t,x,C_{i-1}\right)\right],\\
\delta_{1}\left(t,x,c_{1}\right) & \leq K\left|w_{1}\left(t,c_{1}\right)-\bar{w}_{1}\left(c_{1}\right)\right|.
\end{align*}
This thus gives:
\begin{align*}
 & \mathbb{E}_{Z}\left[\left|{\cal H}\left(X,W\left(t\right)\right)-{\cal H}\left(X,\left\{ \bar{w}_{i}\right\} _{i\leq L}\right)\right|\right]\\
 & \leq K\mathbb{E}_{Z}\left[\delta_{L}\left(t,X,1\right)\right]\\
 & \leq K^{L}\sum_{i=2}^{L}\mathbb{E}\left[\left|w_{i}\left(t,C_{i-1},C_{i}\right)-\bar{w}_{i}\left(C_{i-1},C_{i}\right)\right|\sideset{}{_{j=i+1}^{L}}\prod\left|\bar{w}_{j}\left(C_{j-1},C_{j}\right)\right|\right]\\
 & \quad+K^{L}\mathbb{E}\left[\left|w_{1}\left(t,C_{1}\right)-\bar{w}_{1}\left(C_{1}\right)\right|\sideset{}{_{j=2}^{L}}\prod\left|\bar{w}_{j}\left(C_{j-1},C_{j}\right)\right|\right].
\end{align*}
By the convergence assumption, the right-hand side tends to $0$ as
$t\to\infty$. This proves the claim and concludes the proof.
\end{proof}

\section{Connection to large-width neural networks\label{sec:connection}}

Theorem \ref{thm:global-optimum} concerns with the global convergence
of the MF limit. To make the connection with a finite-width neural
network, we recall the neuronal embedding $\left(\Omega,P,\left\{ w_{i}^{0}\right\} _{i\leq L}\right)$,
as well as the following coupling procedure in \cite{nguyen2020rigorous}:
\begin{enumerate}
\item We form the MF limit $W\left(t\right)$ (for $t\in\mathbb{R}_{\geq0}$)
associated with the neuronal ensemble $\left(\Omega,P\right)$ by
setting the initialization $W\left(0\right)$ to $w_{1}\left(0,\cdot\right)=w_{1}^{0}\left(\cdot\right)$,
$w_{i}\left(0,\cdot,\cdot\right)=w_{i}^{0}\left(\cdot,\cdot\right)$
and running the MF ODEs described in Section \ref{subsec:Mean-field-limit}.
\item We independently sample $C_{i}\left(j_{i}\right)\sim P_{i}$ for $i=1,...,L$
and $j_{i}=1,...,n_{i}$. We then form the neural network initialization
$\mathbf{W}\left(0\right)$ with $\mathbf{w}_{1}\left(0,j_{1}\right)=w_{1}^{0}\left(C_{1}\left(j_{1}\right)\right)$
and $\mathbf{w}_{i}\left(0,j_{i-1},j_{i}\right)=w_{i}^{0}\left(C_{i-1}\left(j_{i-1}\right),C_{i}\left(j_{i}\right)\right)$
for $j_{i}\in\left[n_{i}\right]$. We obtain the network's trajectory
$\mathbf{W}\left(k\right)$ for $k\in\mathbb{N}_{\geq0}$ as in Section
\ref{subsec:Three-layer-neural-network}, with the data $z\left(k\right)$
generated independently of $\left\{ C_{i}\left(j_{i}\right)\right\} _{i\leq L}$
and hence $\mathbf{W}\left(0\right)$.
\end{enumerate}
Here in our present context, the neuronal embedding forms the basis
on which the finite-width neural network is realized. Furthermore
the neural network and its MF limit are coupled. One can establish
a result on their connection, showing that the coupled trajectories
are close to each other with high probability, similar to \cite[Theorem 10]{nguyen2020rigorous}.
Together with Theorem \ref{thm:global-optimum}, one can obtain the
following result on the optimization efficiency of the neural network
with SGD:
\begin{cor}
\label{cor:global-optimum-NN}Consider the neural network (\ref{eq:three-layer-nn})
as described by the coupling procedure. Under the same setting as
Theorem \ref{thm:global-optimum}, in Case 1,
\[
\lim_{t\to\infty}\lim_{\left\{ n_{i}\right\} _{i\leq L}}\lim_{\epsilon\to0}\mathbb{E}_{Z}\left[{\cal L}\left(Y,\hat{{\bf y}}\left(X;\mathbf{W}\left(\left\lfloor t/\epsilon\right\rfloor \right)\right)\right)\right]=\inf_{F}\mathscr{L}\left(F\right)=\inf_{\tilde{y}}\mathbb{E}_{Z}\left[{\cal L}\left(Y,\tilde{y}\left(X\right)\right)\right]
\]
in probability, where the limit of the widths is such that $\left(\min\left\{ n_{i}\right\} _{i\leq L-1}\right)^{-1}\log\left(\max\left\{ n_{i}\right\} _{i\leq L}\right)\to0$.
In Case 2, the same holds with the right-hand side being $0$.
\end{cor}

\bibliographystyle{amsalpha}
\bibliography{NN_companion}

\providecommand{\bysame}{\leavevmode\hbox to3em{\hrulefill}\thinspace}
\providecommand{\MR}{\relax\ifhmode\unskip\space\fi MR }
\providecommand{\MRhref}[2]{%
  \href{http://www.ams.org/mathscinet-getitem?mr=#1}{#2}
}
\providecommand{\href}[2]{#2}
\begin{thebibliography}{MMN18}

\bibitem[AOY19]{araujo2019mean}
Dyego Ara{\'u}jo, Roberto~I Oliveira, and Daniel Yukimura, \emph{A mean-field
  limit for certain deep neural networks}, arXiv preprint arXiv:1906.00193
  (2019).

\bibitem[CB18]{chizat2018}
L\'{e}na\"{\i}c Chizat and Francis Bach, \emph{On the global convergence of
  gradient descent for over-parameterized models using optimal transport},
  Advances in Neural Information Processing Systems, 2018, pp.~3040--3050.

\bibitem[MMN18]{mei2018mean}
Song Mei, Andrea Montanari, and Phan-Minh Nguyen, \emph{A mean field view of
  the landscape of two-layers neural networks}, Proceedings of the National
  Academy of Sciences, vol. 115, 2018, pp.~7665--7671.

\bibitem[Ngu19]{nguyen2019mean}
Phan-Minh Nguyen, \emph{Mean field limit of the learning dynamics of multilayer
  neural networks}, arXiv preprint arXiv:1902.02880 (2019).

\bibitem[NP20]{nguyen2020rigorous}
Phan-Minh Nguyen and Huy~Tuan Pham, \emph{A rigorous framework for the mean
  field limit of multilayer neural networks}, arXiv preprint arXiv:2001.11443
  (2020).

\bibitem[RVE18]{rotskoff2018neural}
Grant Rotskoff and Eric Vanden-Eijnden, \emph{Parameters as interacting
  particles: long time convergence and asymptotic error scaling of neural
  networks}, Advances in Neural Information Processing Systems 31, 2018,
  pp.~7146--7155.

\bibitem[SS18]{sirignano2018mean}
Justin Sirignano and Konstantinos Spiliopoulos, \emph{Mean field analysis of
  neural networks}, arXiv preprint arXiv:1805.01053 (2018).

\bibitem[SS19]{sirignano2019mean}
\bysame, \emph{Mean field analysis of deep neural networks}, arXiv preprint
  arXiv:1903.04440 (2019).

\end{thebibliography}

\end{document}